# Epidemic outbreak prediction using machine learning models

Akshara Pramod[1], JS Abhishek[2], Dr. Suganthi K[3]
School of Electronics Engineering, Vellore Institute of Technology Chennai Campus, Chennai, India - 632014

**Abstract**
In today's world, the risk of emerging and re-emerging epidemics have increased. The recent advancement in healthcare technology has made it possible to predict an epidemic outbreak in a region. Early prediction of an epidemic outbreak greatly helps the authorities to be prepared with the necessary medications and logistics required to keep things in control. In this article, we try to predict the epidemic outbreak (influenza, hepatitis and malaria) for the state of New York, USA using machine and deep learning algorithms, and a portal has been created for the same which can alert the authorities and health care organisations of the region in case of an outbreak. The algorithm takes historical data to predict the possible number of cases for 5 weeks into the future. Non-clinical factors like google search trends, social media data and weather data have also been used to predict the probability of an outbreak.

**Introduction**
More than six different influenza pandemics and epidemics have struck in just a century. Every year nearly 500,000 people die due to seasonal influenza and other epidemics. Even with the advancement of technology, particularly in the healthcare industry, it is impossible to prevent an outbreak, but it's possible to be prepared for one. With the help of machine learning, we can now monitor and forecast the expected number of cases of a given disease for a particular region by using meteorological data, social media data, and historical data. This would be extremely useful for the health care centres and pharmacies of a particular region to be prepared in advance and stock up their inventory if needed.

As seen in India during the second wave of COVID-19, due to the suddenness of the outbreak there was an unprecedented demand in healthcare resources from medicines, beds, etc. Such unexpected epidemic or pandemic outbreaks threaten to overwhelm the healthcare system of any region. But knowing the possibility of an outbreak beforehand helps the healthcare system to be prepared in advance, as it gives them enough time to accumulate the necessary items like medicines, oxygen cylinders, etc. With the help of machine learning and the historical incidence data of any disease, it is now possible to anticipate an outbreak up to five weeks in advance. This will help healthcare centres and pharmacies to be prepared in case of an outbreak. It also enables the government authorities to provide restrictive guidelines for the society and ensure human safety during such times.

There has been a lot of research in this domain by various known authors and researchers to find the most optimal solution in order to ensure proper analysis of outbreak situations. Some works also focus on developing an interactive dashboard to allow the user to visualise the outbreak trends. Visualising and analysing the outbreak prediction can be proven to be the most useful step through which we can alert the healthcare industry to make necessary provisions in the times of outbreak. But they majorly lack in the development of an entire system that is capable of monitoring, evaluating and taking necessary steps all together. Although many papers have addressed the issue of developing a system for detecting COVID-19, such systems for epidemic outbreak haven't been paid much attention to.

Our work aims at creating machine learning models that are capable of forecasting the number of cases of a particular disease in a particular region and then generate the probability of an epidemic outbreak using non-clinical data such as social media data and meteorological data. It includes scraping the required data from multiple social media websites and also collecting historical epidemiological data and meteorological data. The non-clinical parameters are combined and put into a classifier model to calculate the probability of an outbreak. There are many classification models available for such tasks. The purpose of this work is to observe the performance of various classification models and deploy the best performing model on the web portal created.

The main motive behind developing this kind of a system is to ensure both prevention and cure. In order to develop the proposed system, deep learning models like LSTM will be used for efficiency, reliability, and accuracy. The LSTM model is trained to predict the number of cases for five weeks in advance using the precedence data. For a more informed and calculated choice of model, experiment is performed using ARIMA time-series model as well. The results of both the models are discussed in the further sections.

After performing all the necessary data modelling, data mining, and deriving important conclusions, the work focuses on developing an entire system that can be hosted as a web portal named preCURE that is cable of displaying forecasts for various diseases for a span of 50 weeks and also evaluate the forecasted cases and then send alert messages to the users in case of an outbreak. The portal is also embedded with features like: enabling the authority to view the latest news related to the disease and giving the authority a list of medicines to be kept ready in case of an outbreak. This entire system can be proven to be a more reliable, effective, and well-structured model of detecting and preventing the spread of epidemics.

The study contains several parts; elaborated analysis of all related works in the second section. The architecture of the system is described in the third section. The fourth section consists of the results and observations from our study. Finally, the last section consists of the conclusion and possible future works.

**Literature Survey**

**Web Based System for Epidemic Forecasting**
Eldon Y. Li et al (2016), proposed an epidemic Prediction Market system (PMS) that is capable of predicting infectious diseases in the city of Taiwan based on the wisdom of crowds that enables the model to improve the accuracy of the proposed system. The prediction market system combines information from various resources and forecasts the results. Knowledge was accumulated from 126 health professionals for a span of 31 weeks to predict for 5 diseases namely, dengue fever, influenza,rate of ILI, and the rate and cases of enterovirus infections. The system successfully predicted trends for three indicators much better than the existing systems. Suruchi Deodhar et al (2015), created FluCaster, a web application for high-res situation assessment and predicting the number of outbreaks for flu. The application helps users to view state and regional level influenza cases using high resolution maps, and also can give forecasts for that particular region. The UI can render the dynamic maps without any delays or performance degradation. Shahriare Satu et al (2021), developed a portal to update and forecast real-time information of COVID-19 cases in Bangladesh. The portal allows users to view relevant information like, the emergency helpline numbers, nearby health centres and pharmacies. The portal also used machine learning models to provide short-term forecasting web tools which is able to predict the number of cases for upto 2 days in the future. RMSE, MAE and $R^2$ values were used to evaluate the performances of different models. Zhenfeng Lei et al (2019), developed a d-DC, a data-driven framework, which collects a combination of clinical data like personal or family medical records and uses traditional methods to develop a data acquisition model. The model classifies the disease using machine learning techniques and knowledge graphs (KG), which in particular classifies the diseases initially. In this work a fusion method RKRE is applied which uses ResNet. The fusion method provides an average proportion of 86.5%. Maldi Salehi et al (2021), has worked on developing a useful interactive dashboard that enables real-time monitoring of COVID-19 cases and visualising them on the dashboard. The dashboard provides data in 5 different dimensions of interest namely: time-series plot, datasets, demographics, growth modelling and spatial analysis. The dashboard was developed as a cross-platform web browser. The final observations were quite promising and accurate. The overall work consisted of statistical analysis deployed in the form of visualisation for quick and efficient analysis and precise monitoring. Hemant Ghayvat et al (2021), developed a system named COUNTERACT that involves big data and analysis of the same for identifying COVID-19 hotspots and outbreak predictions and displaying them on the application. The technologies used in this work for prediction are Internet of Things, Edge Computing and Artificial Intelligence. The system uses suspicion modelling, behavioural modelling and backpropagation methods, and LSTM on the spatiotemporal data. The ROC curves were used to visualise the model's accuracy with the complete database and different percentages of missing entries in the database. The data with complete entities showed exceptional accuracy and performance scores. Mitra et al (2021), in his work has stated the proof of concept regarding a real-time monitoring system for COVID-19 and outbreak prediction using epidemiological data. The work includes an interactive dashboard used to visualise the results. The epidemiological data for the first wave is taken to forecast the onset of the second wave in India. The epidemic curves are used for visualisation. The time-dependent reproduction(Rt) has been calculated and various observations are made by percentage reduction of Rt to plot the daily incidences.

**Deep Learning based Methods**
Mengyang Wang et al (2019), used a stacking architecture to predict the number of incidences of malaria as the traditional time series models are ineffective in capturing every single detail present in the data structure.. They used a stacking architecture that can combine distinct algorithms and models. The ARIMA, BP-ANN, STL+ARIMA, and LSTM network models were separately implemented on malaria and meteorological data. The performance was compared using evaluation metrics like RMSE, MAD, and MASE. The RMSE, MAD, and MASE values decreased after the use of stacking architecture. Yuexin Wu et al (2018), for the first time, used deep learning methods for real-time prediction. They used RNN to capture the long-term correlation in the data and CNNs to fuse information from many different sources. Their methods provided better results than the usual baseline autoregressive and Gaussian process models. Leili Tapak et al (2019), investigated the accuracy provided by random-forest time series, support vector machine, and artificial neural network models in ILI modelling and detection of disease outbreaks. The weekly ILI frequencies dataset of Iran was used for prediction. As a result, it was concluded that the random-forest time series model outperformed the other three models in modelling the data. In addition, it was found that the neural network model was better in outbreaks detection. Marwah Soliman et al (2019), studied seasonal influenza trends in Dallas County. The work aims at assessing the forecasting capability of different deep learning models with the use of feedforward neural networks and assessing the performance of conventional statistical models. This work includes the development of a probabilistic forecasting model based on the influenza morbidity information in the country by fusing all the models with the help of Bayesian model averaging. The results suggest that DL and BMA-based multi-model ensemble of ILI forecasts yield almost similar performance which outperforms all the other models tested. Wenxiao Jia et al (2019), created prediction models based on mortality incidence and historical morbidity data of infectious diseases. They also integrated search engine query data and seasonal information for more enhanced results. The different models used for forecasting the morbidity of 10 infectious diseases including the linear model, boosting tree model, time series analysis, and deep learning models (RNNs) were constructed for prediction purposes. It was found that the RNN model outperforms all the other prediction models. Abdul Mahatir Najar et al (2018), predicted the risk level of Dengue Hemorrhagic Fever (DHF) outbreak using Extreme Learning Machines (ELM). It was established that the spread of DHF is connected to the climate conditions of the region. The ELM model takes weather data as input and risk level of DHF outbreak as the target. The ELM model with 50 hidden neurons and binary sigmoid activation function performs best when compared to other ELM models used. Soheila Molaei et al (2019), employed non-linear methods like ARX, ARMAX, Deep Multilayer Perceptron and a convolutional neural network (CNN), to predict influenza incidence based on social media data like twitter. The product of the tweets and Centres for Disease Control and Prevention (CDC) data and of the tweets and google data were also included as new features. The deep neural network trained on the newly added features performed better as compared to the other methods. The model reduced the mean average error by upto 25%.

**Machine Learning based Methods**
Mauricio Santillana et al (2015), proposed a methodology completely based on machine learning that provides real-time forecasts of influenza activity in the US by collecting data from multiple data sources both clinical and non-clinical. Their system was able to forecast up to the next four weeks in accordance with the CDC's ILI reports. Yirong Chen et al (2018), assessed how the LASSO machine learning model may be helpful in providing useful prediction results for a given input of dissimilar pathogens in areas with varied climates. A set of LASSO models were created each for different diseases, different countries, and different forecasting windows with variable model complexities. Prediction models were nearly accurate in predicting the outbreak but were insensitive while capturing the size of outbreak. Achyuth Ajith et al (2020), studied various models used in predicting the epidemic outbreak caused by the West Nile Virus (WNV). They also discuss the use of different compartmental models in determining the spread of the West Nile Virus. It was concluded that the Random Forest Classifier showed better performance for predictions. A Susceptible Infected Susceptible (SIS) was able to present the best spread model. Ruiri Liang et al (2019), constructed a prediction model for predicting the outbreak of African Swine Fever (ASF) using ASF outbreak data and the meteorological data. Random Forest algorithms were used to construct the model. The model achieved an accuracy range of 76.02% - 84.64%. The work does not include all terrain related data and use of other feature selection techniques and deep learning techniques could have provided better results. Fangge Li et al (2011), showed how ARMA models can be used to forecast the incidence of NewCastle disease for a span of one month in a province in China. The ARMA model with 0 auto regressive orders and 1 moving average order was able to provide accurate and reliable forecasts of New Castle outbreaks. The parameters needed for prediction were estimated using SPSS. Weekly or daily data could have provided better results as compared to monthly data. Agranee Jha et al (2020), predicted the possibility of malaria outbreak in India using machine learning algorithms like Support Vector Machine (SVM). The climate data like Rainfall, Temperature and Humidity is used to predict the outbreak. The output parameter is a binary class of Yes (malaria outbreak) or No (no malaria outbreak). The model could have provided better results had it also considered other clinical or non-clinical parameters like social media data etc.

Numerous research works were performed using various machine learning and deep learning techniques separately. There are a few interactive dashboard systems that were implemented for the user to visualise the data and analyse the epidemic/pandemic situations. However, there is a need for a full-fledged monitoring and prevention system. Therefore, the current work aims at developing a system that integrates the process of monitoring the chances of occurrence of the epidemic and the task of alerting the responsible authorities in order to overcome the shortcomings of the previous works implemented in the same domain.

**Proposed System**
**Proposed Architecture**
The proposed architecture is a combination of monitoring and alerting components. The system uses non-clinical parameters and various machine learning and deep learning technologies for epidemic outbreak prediction. The client-side architecture of the system involves mainly two entities; the admin ( any medical superior or government authority) and the users (healthcare centres, pharmacies or hospitals). The admin monitors the previous number of cases of mainly three epidemics like malaria, hepatitis and influenza which spread frequently in an area like New York. In addition to the past cases, the admin can also analyse the situation based on the forecasted number of cases for the next five weeks. To keep the admin more informed, the system includes a feature of displaying the news related to the three pandemics covered in this work for a span of 7 days. The proposed model allows the admin to evaluate the situation and as per the situation send alert messages to all the registered pharmacies, healthcare centres and hospitals to stock up their inventory and prepare themselves for the time of emergency. Figure 1 Client Side Architecture shows a precise description of the client-side architecture.

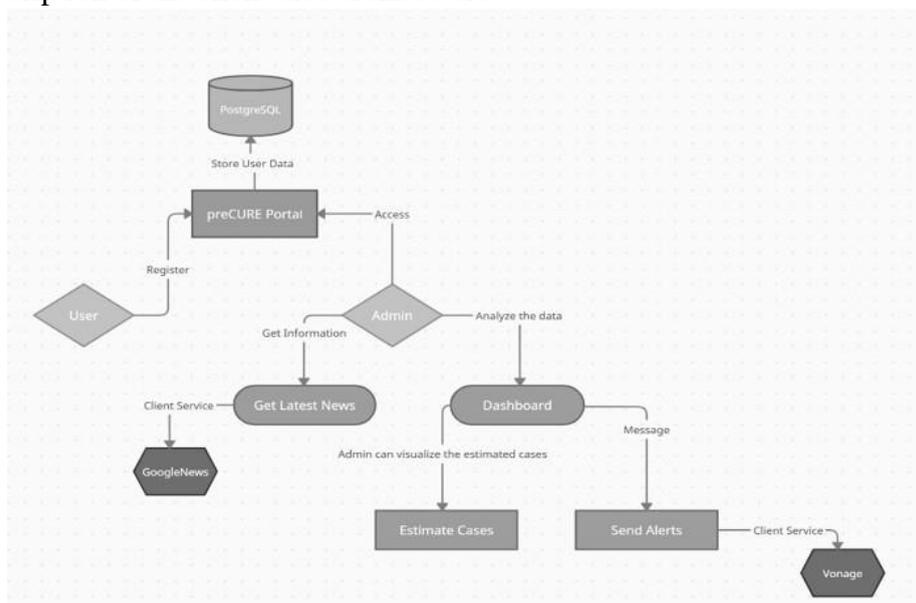

**Figure 1** Client Side Architecture

On the contrary the server-side architecture is responsible for collecting the weekly data, shaping the data into the required format,and preprocessing the data in the initial stage. Once the data is collected and formatted it is merged to create a wholesome dataset with four parameters like weekly average temperature, weekly average precipitation, weekly count of google search trends, and weekly count of twitter data. The historical epidemiological data is set as the target parameter. The dataset is preprocessed to overcome data discrepancies and the probability of occurrence of an epidemic is calculated using nine different machine learning models like, Naive Bayes, Decision Tree, Logistic Regression, KNN, Support Vector Machine, Bagging and Boosting Decision Trees, Voting Classifier, and Neural Networks, and the model giving the best performance is deployed in the web portal. Similarly, the epidemiological data is trained and tested using two time-series forecasting methods, namely, ARIMA and LSTM. The number of cases are forecasted for next five weeks, and the time-series model with better results is deployed on the web application for evaluation. Figure 2 Server Side Architecture explains the complete server-side flow.

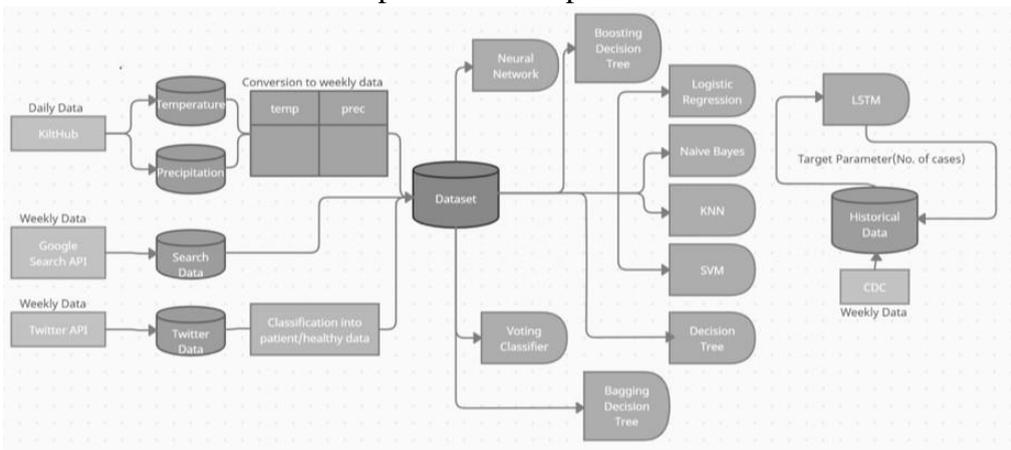

**Figure 2** Server Side Architecture

The elaborated methodology to explain how the system performs all the tasks will be discussed in this section ahead.

**Data Source**
The proposed system needs a combination of various non-clinical parameters for a span of 5 years ranging from January 2017 to December 2021 to enhance the efficiency and accuracy of prediction models. The data is collected using various resources :
- Firstly, the weather data that constitutes daily temperature and daily precipitation in the state of New York for a span of five years was collected from Kilthub, a source provided by Carnegie Mellon University.
  The required format of data is weekly format, therefore, the daily data is converted into weekly data using pivot table method of MS Excel. During the time of implementation of the web portal, the weather data is collected using the meteostat library in python for automated values.
- Secondly, the google search volume data is collected on a weekly basis using Google Search API ( pytrends) for the state of New York. The same method is used for automating and directly inserting the further volume data into the dataset during the execution of the portal.
- Thirdly, the twitter count data is collected on a weekly basis for a span of five years using Twitter API (tweepy). It uses keywords like malaria, hepatitis, flu, and influenza. The influenza count is collected as a sum of flu count and influenza count.The data for automation during the execution process is collected using sntwitter python library.
- Lastly, the historical epidemiological data that depicts the total number of cases of influenza, malaria and hepatitis in the state of New York is collected from the Centre for Disease Control and Prevention. The future data is inserted into the database by the authority manually during the process of evaluation.

The entire data is collected, sorted, and cleaned and then further converted into a single dataset which is used in the proposed system for all the training and testing purposes.

**Data Preprocessing(Min-Max Scaling)**
Min-Max scaling or min-max normalisation is a method for data normalisation. The range of features are rescaled to be in the range of [0, 1] or [-1, 1]. The general formula for a min-max of [0, 1] is given as:

$$x' = x - min(x) / (max(x) - min(x))$$

here $x$ is an original value and $x'$ is the normalised value. The MinMaxScaler from the sklearn library was used to normalise the data.

**Time-Series Forecasting**

Time series forecasting is a method in which we can make predictions based on time-stamped historical data. The historical data is analysed and the observations made from them are used to make decisions. Two most commonly used Time series methods are ARIMA and LSTM (deep learning). Following subsections describe briefly about ARIMA and LSTM models.

**1. Auto-Regressive Integrated Moving Average (ARIMA)**

ARIMA is a time series method used to understand the data and predict future values in the series. All non-seasonal data could be modelled with ARIMA. An ARIMA model has three terms namely, p, d, and q. Here p is the order of the Auto Regressive term, d is the order of differencing and q is the order of moving average term.

The Auto-Regressive model uses its own lag to forecast the future values. Linear regression models work best when the independent features have no correlation with each other. Hence it's important to make the data stationary. The easiest way to do so is by differencing. Hence d determines the minimum number of differences needed to make the data stationary. The moving Average model uses errors to forecast the future values.

$$Y_t = \alpha + \beta_1 Y_{t-1} + \beta_2 Y_{t-2} + .... + \beta_p Y_{t-p} + \epsilon_1 \qquad (1)$$
$$Y_t = \alpha + \epsilon_t + \phi_1 \epsilon_{t-1} + \phi_2 \epsilon_{t-2} + .... + \phi_q \epsilon_{t-q} \qquad (2)$$
$$Y_t = \alpha + \beta_1 Y_{t-1} + \beta_2 Y_{t-2} + .... + \beta_p Y_{t-p} \epsilon_t + \phi_1 \epsilon_{t-1} + \phi_2 \epsilon_{t-2} + .... + \phi_q \epsilon_{t-q}$$

Here equation (1) describes the autoregressive model and (2) the moving average model. The final equation represents the ARIMA model. Here, β and ϕ are the coefficients, and ϵ term represents the errors of the AR models of the respective lags.

auto_arima from the pmdarima library of python was used to find the best-fit parameters of the ARIMA model. Once the best parameters are found, then the ARIMA model from the statsmodel package is used to train the model.

**2. Long Short Term Memory Model**

Long Short-Term Memory is an artificial recurrent neural network (RNN) used in Deep Learning. Unlike the traditional feedforward neural networks, RNNs with feedback loops in them allow the information to persist. This enables them to connect previous information to the present task.

LSTMs are a special type of RNN that can remember information for a longer period of time. Each LSTM cell has an input gate, output gate, and forget gate. These gates are the computing units that control the activation of the cell. The forget gate multiplies the previous state of the cell while the input and output gates multiply the input and output of the cell. Logistic sigmoid is used as the activation function. As long as the forget gate is open and the input gate is closed, the memory cell continues to remember the first input.

The equations for forgetting, storing, renewing, and outputting information in the cell are shown below, respectively.

$$f_t = \sigma(w_f \cdot [h_{t-1}, x_t] + b_f) \quad (3)$$
$$i_t = \sigma(W_i \cdot [h_{t-1}, x_t] + b_i) \quad (4)$$
$$\tilde{C}_t = \tanh(W_c \cdot [h_{t-1}, x_t] + b_C) \quad (5)$$
$$C_t = f_t \times C_{t-1} + i_t \times \tilde{C}_t \quad (6)$$
$$o_t = \sigma(W_o \cdot [h_{t-1}, x_t] + b_o) \quad (7)$$
$$h_t = o_t \times \tanh(C_t) \quad (8)$$

When data ($x_t$) is input to the LSTM cell in equation (3), the function $f_t$ determines the information to be forgotten in the cell layer. In equations (4) and (5), information that will be newly saved in the cell layer is created in $i_t$ and $\tilde{C}_t$. In equation (6), the cell layer $C_t$ is renewed using $f_t$, $i_t$, and $\tilde{C}_t$. In equation (7), the cell layer's information is used and $h_t$ is the output. In equation (8), the cell state gets a value between -1 to 1 through tanh function. The values of $C_t$ and $h_t$ are kept for the next iteration of LSTM.

**Methodology for Time Series Forecasting:**
The methodology for time series forecasting used is mentioned below:
1. The historical morbidity data for influenza, malaria and hepatitis were collected from the Centre for Disease Control and Prevention over the time period of five years (2017-2021) for the state of New York, USA.
2. The collected data was then preprocessed and normalised, as the time series data has input values with different scales.
3. The data was then used to train the LSTM time series models.
4. The model with the lowest RMSE score and the one which could capture the seasonality and the variations in the data was used to make the final forecast.

Based on previous research it was found that LSTM models outperformed traditional models like ARIMA and were better at capturing the seasonal trends in the data. Hence LSTM models were used for time series forecasting. Three LSTM models were developed, one for each disease. The model developed for Influenza and Hepatitis, has two LSTM layers each with 64 and 32 units respectively. The output of the LSTM layer is connected to a dense layer with 32 neurons. The output from this layer is connected to a final output layer. Mean Squared error was used as the loss function and adam optimizer was used to optimise the weights of the network. The model developed for malaria has two LSTM layers each with 128 and 64 units respectively. The output of LSTM layers is connected to a dense layer with 64 neurons. The output of this layer is connected to a final output layer.

**Probability of Occurrence**
Various research has clearly shown the use of non-medical data like meteorological data, and social media data in epidemic outbreak prediction. Here, we have taken parameters like weekly precipitation, temperature, google search trends, and tweet count to predict the probability of an epidemic outbreak. The historical incidence data has been converted into a categorical data (0 for no epidemic, 1 for epidemic) based on the criteria that if the number of cases for a week exceeds the mean incidence value of the last 5 years, then it can be considered an outbreak as in such cases the healthcare system would require more resources to accommodate and treat the patients. The dataset was then trained using various supervised classification algorithms like, Logistic Regression, Naive Bayes Classification, KNN,

Decision Tree based algorithms and random forest classifier. The model with the best performance (accuracy) was used for final deployment.

**Methodology for Probability Calculation**
Methodology for probability classification:
1. Three datasets were prepared each containing parameters like Weekly Precipitation, Temperature, Google Search Trends, Tweet count of the current week, and epidemic calculation of the next week.
2. The data was normalised when necessary using Min-Max Scaler and Standard Scaler.
3. Each of the datasets was trained with different classification models namely, Logistic Regression, Naive Bayes Classifier, K Nearest Neighbour, SVM, Decision Tree (Standard, Bagging and Boosting), Random Forest, Voting Classifier, and a feed-forward neural network.
4. The models which performed the best were used for the final deployment.

**Web Application**
During the development of the preCURE portal many libraries, clients, softwares, algorithms, and languages were used.

The application uses the gnewsclient to display the weekly news regarding each disease and the vonage api client to send alert messages. Various libraries that support flask were used to build the application. The reason behind using flask for this application is that it is light-weight and also supports numerous functionalities. Table 1, 2 gives a brief description about the languages and software specifications used while building this model.

Table 1       Languages Used

| Name of Language | Description and Usage |
|---|---|
| **Python** | The Flask web framework is used to develop this application. Flask is written in Python and is easy to read. The Machine Learning model has also been written in Python due to the numerous features and libraries provided by Python and its robustness. |
| **HTML** | The views i.e. the client-side web application's structure is coded in HTML. HTML is a widely used language, and is easy to use. It is also supported by every browser. |
| **CSS** | CSS provides better website speed and is easy to maintain. It is used for styling the structure that is written in HTML to enhance the view of the website. |

Table 2       Softwares Used

| Name of Software | Description and Usage |
|---|---|
| **Google Colaboratory** | Google Colab is used to train the model and convert it into pickle files that are directly used in the application later for forecasting. |
| **MS Excel** | MS Excel is used to convert the daily weather data to weekly data with the use of Pivot Table Method. |

The database used in the preCURE portal is a Heroku PostgreSQL. PostgreSQL is a simple Relational database with a well defined schema. The database is implemented with the help of SQLAlchemy framework and psycopg2 connection is used to execute the SQL commands in raw SQL language. The application doesn't require any complex structuring or querying hence, relational databases can be used.

**Schema:**
- **Admin Table:** consists of id, admin's name, email and password. During the login the login manager checks if the admin with this email and password exists and based on that it logs in the admin.
- **Users Table:** consists of user's name, phone number, organisation name, organisation address, category and email id. The phone numbers are used to send alerts to the registered users.
- **Malaria Table:** consists of id, weekly precipitation, weekly temperature, weekly google trends data, weekly twitter count, and the incidence data of malaria.
- **Hepatitis Table:** consists of id, weekly precipitation, weekly temperature, weekly google trends data, weekly twitter count, and the incidence data of hepatitis.
- **Influenza Table:** consists of id, weekly precipitation, weekly temperature, weekly google trends data, weekly twitter count, and the incidence data of influenza.

The preCURE portal provides certain functionalities that are as follows:
- **Extracting Data from the SQL:** This function is used to extract the last 50 weeks data from the SQL which is further plotted on the website using the plot function.
- **Forecasting the Number of Cases:** This function is used to forecast the number of cases for the next five weeks based on the previous cases extracted from the previous function and the pretrained model is used to predict the future cases.
- **Plotting the Data:** The previous cases and forecasted cases are visualised using a plotly plot in this function and then sent as a json object to the HTML file for visualisation in the website.
- **Classification Probability:** This function calculates the probability of occurrence of the epidemic based on non-clinical parameters.
- **Extracting the tweet count:** This function uses the sntwitter module to scrape the tweet count of three diseases based on the start date and the end date.
- **Extracting the Google Search Volume Data:** This function uses the pytrends module to scrape the weekly google search volume data of malaria, hepatitis, and influenza.
- **Extracting the Weather Parameter:** The meteostat module is used in this function to scrape the open weather parameters.
- **Insertion into SQL:** This function inserts the id, weather parameters, google search data, twitter count, and the number of epidemic cases in the SQL tables. The epidemic cases are taken as an input from the admin.

## Results

### Time-Series Forecasting

For time series forecasting, we found that the LSTM model performed much better than the conventional ARIMA model. Figure 3 and Figure 4 depicts that the LSTM model was able to capture the variations and seasonality in the data as compared to the ARIMA model.

Table 3          ARIMA Model Results

| Disease | ARIMA Best Model (p, d, q) | RMSE |
|---|---|---|
| Hepatitis | (0, 1, 2) | 0.50377 |
| Malaria | (0, 1, 1) | 1.48759 |
| Influenza | (1, 1, 3) | 212.5937 |

The mean of the test dataset used for the ARIMA prediction is similar to the RMSE of the model, indicating that the ARIMA models could not learn any seasonal trends in the data. LSTM models on the other hand were able to learn the variations in the data and were able to give closer to real life results. All the LSTM model variants tried had two LSTM layers, as the LSTM model with two layers was able to learn the variations much better than LSTM models with one LSTM layer. The sequence size of these models was set as 5, i.e the model took the previous 5-week data to forecast the next week's data. Mean squared error was used as the loss function, and adam function was used to optimise the learning parameters of the model. The models were trained for 100 epochs each. The results of the models are shown below.

Table 4          LSTM Model Results

| Influenza | | Malaria | | Hepatitis | |
|---|---|---|---|---|---|
| LSTM Units | RMSE | LSTM Units | RMSE | LSTM Units | RMSE |
| 128-64 | 32.35 | **128-64** | **1.1** | 128-64 | 3.47 |
| **64-32** | **29.57** | 64-32 | 1.2 | **64-32** | **3.16** |
| 32-16 | 31.26 | 32-16 | 1.15 | 32-16 | 3.18 |

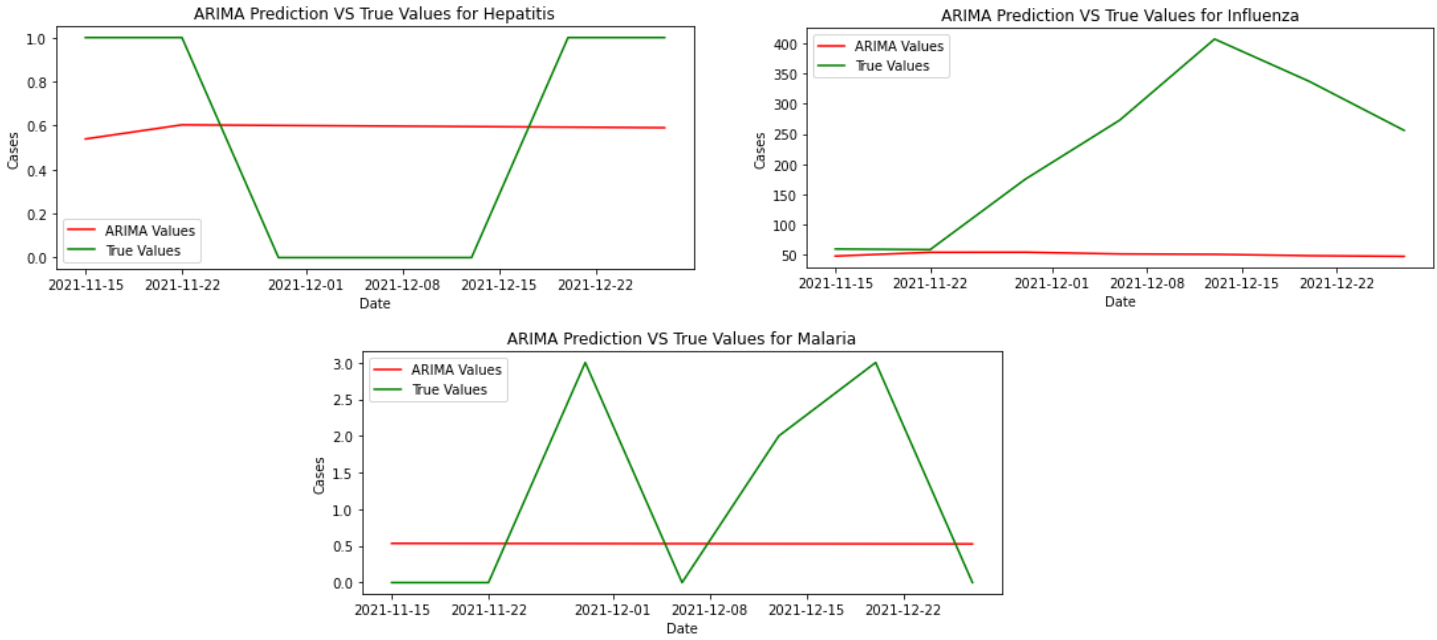

**Figure 3** ARIMA model results for each of the three diseases

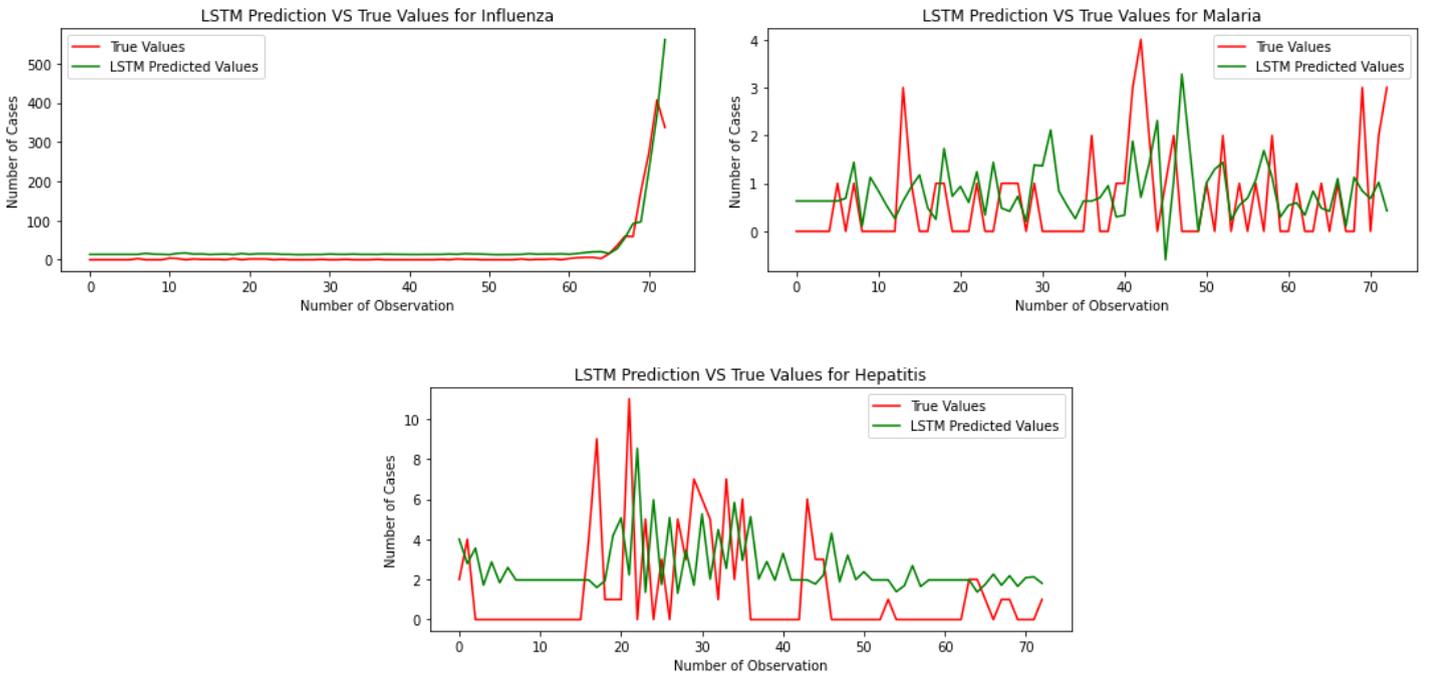

**Figure 4** LSTM model results for each of the three diseases

**Outbreak Probability**

For Probability classification, the decision tree-based classifiers like bagging classifier and random forest performed better than the other classification models. The dataset was trained with different classification models like Logistic Regression, Naive Bayes, K Nearest Neighbours, SVM, Decision Tree-Based Classifiers and a feed-forward neural network. The results of the following are given in the table below.

**Table 5** Probability Classification Results

| Influenza | | Malaria | | Hepatitis | |
|---|---|---|---|---|---|
| **Model** | **Accuracy** | **Model** | **Accuracy** | **Model** | **Accuracy** |
| Logistic Regression | 87.69 | Logistic Regression | 89.23 | Logistic Regression | 53.84 |
| Naive Bayes | 73.84 | Naive Bayes | 89.23 | Naive Bayes | 58.46 |
| KNN | 86.15 | KNN | 89.23 | KNN | 52.30 |
| SVM | 75.38 | SVM | 89.23 | SVM | 58.46 |
| Decision Tree Clf. | 90.01 (overfitting) | Decision Tree Clf. | 84.61 (overfitting) | Decision Tree Clf. | 53.84 (overfitting) |
| Bagging D.Tree Clf. | 89.23 | **Bagging D.Tree Clf.** | **89.23** | Bagging D.Tree Clf. | 53.84 |
| Boosting D. Tree Clf. | 90.76 (overfitting) | Boosting D. Tree Clf. | 83.07 (overfitting) | Boosting D. Tree Clf. | 52.30 |
| **Random Forest** | **89.23** | Random Forest | 84.61 | **Random Forest** | **55.76** |
| Voting Clf. | 87.69 | Voting Clf. | 89.23 | Voting Clf. | 53.84 |
| Neural Network | 84.6 | Neural Network | 83.7 | Neural Network | 55.4 |

Some of the traditional models like Logistic Regression, SVM, KNN, and Naive Bayes despite having higher accuracy have learned to predict the same optimal output regardless of the input. Hence we have used random forest and bagging decision tree models which have good accuracy and do not predict the same optimal output.

**preCURE Portal**

The preCURE portal has several functionalities like displaying the informative content for admin like the weekly news regarding the three particular diseases, displaying the forecasts on the website, and sending alerts to healthcare organisations based on the forecasting evaluation. It allows healthcare organisations to register as a preCURE member so that they can get timely alerts in the times of risk via the admin and stock up their inventory.

The NEWS headline page enables the users to read the latest news related to the disease. This allows the users to be updated of the new information like latest medicines discovered or the disease infection report etc. As seen in the image, the google NEWS client provides the latest headlines for the three diseases.

Weekly forecast dashboards, allows the admin to monitor the trends in malaria/hepatitis/influenza infections and also view the forecast for the next 5 weeks. Based on the forecast the admin can then decide whether to send alerts or not. The dashboard also allows the admin to view the necessary medications needed for the disease. It also displays the probability of an epidemic based on non-clinical parameters like temperature, precipitation, google search trends and tweet counts.

This portal allows the admin to send alerts. The admin can send alerts either for a particular disease or all of them. The alerts can also be sent either to the pharmacies or to the health centres or to the hospitals or to all of them. This allows the health organisation to be prepared with the necessary logistics required for the treatment well in advance.

**Discussions**

In this study, we have created a web portal and alert system which would alert the health care organisations in case of an epidemic outbreak. The outbreak forecasting is done based on the historical incidence data of the particular region. Additionally the non-clinical parameters were also used to predict the probability of an epidemic outbreak. The study yielded some key findings like the supremacy of LSTM models in time series forecasting when compared to traditional models like ARMA, and ARIMA. Our results are in accordance with Wenxioa Jia (2019), Yuexin Wu (2018), Mengyang Wang (2019), which also proved that the LSTM time-series model outperforms other traditional time-series models. With regards to the probability prediction, Random Forest models and Bagging Tree models were able to give the best accuracy. Given the non-linear nature of the non-clinical data, the tree based models performed better when compared to the traditional linear models. Our results are in line with the results of Achyith Ajith (2020), which proved that the random forest models gave better accuracy for West Nile Virus detection when compared to models like Adaptive boost, Naive Bayes etc. Hence, this system is able to send forecasts for upto 5 weeks into the future, which helps the health care organisations like pharmaceuticals and hospitals to be prepared in advance.

**Conclusion**

To conclude, this work discusses the development of an entire portal that was developed in order to provide a solution for one of the most important problem statements in the field of healthcare. The purpose was to suggest an entire healthcare system that focuses at observing the previous trends and allowing the concerned authority to take necessary actions to predict and prevent the outbreak of epidemic in their area. In this work, we have focused on majorly three concepts: firstly, forecasting the number of cases of three different epidemics in the state of New York based on the previous five years data; secondly, calculating the probability of occurrence of the epidemic in that particular area based on non-clinical parameters like weather and social media data; lastly, developing a web-based application to allow authorities to monitor the number of cases and alert the concerned health care centres to stock up their inventory. All the three major tasks were performed separately and were merged to develop the entire system. The results suggest that this model can prove to be very reliable in terms of epidemic prediction and also displays good accuracy in accordance with the existing data.

In future, with the help of advanced time series models, it is possible to create a more complex model which would be able to predict an outbreak more accurately. Based on the previous health records and the population of a region, we would be able to calculate the number of medicines and logistics needed to treat a patient in case of an outbreak.